\documentclass[journal]{IEEEtran}

\pdfoutput=1

\ifCLASSINFOpdf

\else

\fi

\usepackage{amssymb}
\usepackage{amsthm}
\usepackage{amsmath}
\usepackage{algorithm}
\usepackage{algpseudocode}
\usepackage{graphicx}
\usepackage{xfrac}
\usepackage{caption}
\usepackage{mathtools}
\usepackage{array}
\usepackage{cite, bm}
\usepackage{subcaption}
\usepackage{multirow}
\usepackage{longtable}
\usepackage{float}
\usepackage[outdir=./]{epstopdf}
\usepackage[export]{adjustbox}
\usepackage{soul,color}
\usepackage{hyperref}
\usepackage{fancyhdr}
\usepackage{makebox}

\newcommand\norm[1]{\left\lVert#1\right\rVert}

\graphicspath{{figures/}}
\usepackage[absolute,showboxes]{textpos}

\TPMargin{5pt}

\newcommand{\copyrightstatement}{
    \begin{textblock}{0.84}(0.08,0.03)    
         \noindent
         \footnotesize
         \copyright2017 IEEE. Personal use of this material is permitted. Permission from IEEE must be obtained for all other uses. Cite: B.O. Ayinde and J. M. Zurada," Deep Learning of Constrained Autoencoders for Enhanced Understanding of Data" in IEEE Trans. on Neural Networks and Learning Systems, September 2018, Vol. 29, Issue 9, Pg. 3969 - 3979.
    \end{textblock}
}

\begin{document}
\copyrightstatement
\title{Deep Learning of Nonnegativity-Constrained Autoencoders for Enhanced Understanding of Data}

\author{Babajide~O. Ayinde,~\IEEEmembership{Student~Member,~IEEE}, and
	Jacek~M.~Zurada,~\IEEEmembership{Life~Fellow,~IEEE}
	\thanks{B.~O.~Ayinde is with the Department
		of Electrical and Computer Engineering, University of Louisville, Louisville,
		KY, 40292 USA (e-mail: babajide.ayinde@louisville.edu).}
	\thanks{J.~M.~Zurada is with the Department
		of Electrical and Computer Engineering, University of Louisville, Louisville,
		KY, 40292 USA, and also with the Information Technology Institute, University of Social Science,\L \'{o}dz 90-113, Poland (Corresponding author, e-mail: jacek.zurada@louisville.edu).}
    \thanks{This work was supported in part by the NSF under grant 1641042.}
}

\maketitle

\begin{abstract}
Unsupervised feature extractors are known to perform an efficient and discriminative representation of data. Insight into the mappings they perform and human ability to understand them, however, remain very limited. This is especially prominent when multilayer deep learning architectures are used. This paper demonstrates how to remove these bottlenecks within the architecture of Nonnegativity Constrained Autoencoder (NCSAE). It is shown that by using both L1 and L2 regularization that induce nonnegativity of weights, most of the weights in the network become constrained to be nonnegative thereby resulting into a more understandable structure with minute deterioration in classification accuracy. Also, this proposed approach extracts features that are more sparse and produces additional output layer sparsification. The method is analyzed for accuracy and feature interpretation on the MNIST data, the NORB normalized uniform object data, and the Reuters text categorization dataset.
\end{abstract}


\IEEEpeerreviewmaketitle

\section{Introduction}
\everymath{\displaystyle}
\IEEEPARstart{D}{eep} learning (DL) networks take the form of heuristic and rich architectures that develop unique intermediate data representation. The complexity of architectures is reflected by both the sizes of layers and, for a large number of data sets reported in the literature, also by the processing. In fact, the architectural complexity and the excessive number of weights and units are often built in into the DL data representation by design and are deliberate \cite{bengio2007scaling,bengio2009learning,Hinton2006reducing,Deng2014tutorial,Bengio2013Guest}. Although deep architectures are capable of learning highly complex mappings, they are difficult to train,  and it is usually hard to interpret what each layer has learnt. Moreover, gradient-based optimization with random initialization used in training is susceptible to converging to local minima \cite{Bengio2007Greedy,ayinde2016clustering}.\\
\indent
In addition, it is generally believed that humans analyze complex interactions by breaking them into isolated and understandable hierarchical concepts. The emergence of part-based representation in human cognition can be conceptually tied to the nonnegativity constraints \cite{lee1999learning}. One way to enable easier human understandability of concepts in neural networks is to constrain the network's weights to be nonnegative. Note that such representation through nonnegative weights of a multilayer network perceptron can implement any shattering of points provided suitable negative bias values are used \cite{Chorowski2014Learning}.\\
\indent
Drawing inspiration from the idea of Nonnegative Matrix Factorization (NMF) and sparse coding \cite{Olshausen1996Emergence,lee1999learning}, the hidden structure of data can be unfolded by learning features that have capabilities to model the data in parts. Although NMF enforces the encoding of both the data and features to be nonnegative thereby resulting in additive data representation, however, incorporating sparse coding within NMF for the purpose of encoding data is computationally expensive, while with AEs, this incorporation is learning-based and fast. In addition, the performance of a deep network can be enhanced using Nonnegativity Constrained Sparse Autoencoder (NCAE) with part-based data representation capability \cite{Ehsan2015Deep,Ranzato2007SparseFeature}. \\
\indent
It is remarked that weight regularization is a concept that has been employed both in the understandability and generalization context. It is used to suppress magnitudes of the weights by reducing the sum of their squares. Enhancement in sparsity can also be achieved by penalizing sum of absolute values of the weights rather than the sum of their squares \cite{ishikawa1996structural,bartlett1998sample,gnecco2010regularization,moody1995simple,ogundijo2017reverse}. In this paper, the work proposed in \cite{Ehsan2015Deep} is extended by modifying the cost function to extract more sparse features, encouraging nonnegativity of the network weights, and enhancing the understandability of the data. Other related model is the Nonnegative Sparse Autoencoder (NNSAE) trained with an online algorithm with tied weights and linear output activation function to mitigate the training hassle \cite{Lemme2012OnlineLearning}. While \cite{Lemme2012OnlineLearning} uses a piecewise linear decay function to enforce nonnegativity and focuses on shallow architecture, the proposed uses a composite norm with focus on deep architectures. Dropout is another recently introduced and widely used heuristic to sparsify AEs and prevent overfitting by randomly dropping units and their connections from the neural network during training \cite{hinton2012improving,srivastava2014dropout}. \\
\indent
More recently, different paradigm of AEs that constrain the output of encoder to follow a chosen prior distribution have been proposed \cite{kingma2013auto,makhzani2015adversarial,burda2015importance}. In variational autoencoding, the decoder is trained to reconstruct the input from samples that follow chosen prior using variational inference  \cite{kingma2013auto}. Realistic data points can be reconstructed in the original data space by feeding the decoder with samples from chosen prior distribution. On the other hand, adversarial AE matches the encoder's output distribution to an arbitrary prior distribution using adversarial training with discriminator and the generator \cite{makhzani2015adversarial}. Upon adversarial training, encoder learns to map data distribution to the prior distribution.\\
\indent
The problem addressed here is three-fold: (i) The interpretability of AE-based deep layer architecture fostered by enforcing high degree of weight's nonnegativity in the network. This improves on NCAEs that show negative weights despite imposing nonnegativity constraints on the network's weights \cite{Ehsan2015Deep}.  (ii) It is demonstrated how the proposed architecture can be utilized to extract meaningful representations that unearth the hidden structure of a high-dimensional data. (iii) It is shown that the resulting nonnegative AEs do not deteriorate their classification performance. This paper considerably expands the scope of the AE model first introduced in \cite{baba101} by: (i) introducing smoothing function for $L_1$ regularization for numerical stability, (ii) illustrating the connection between the proposed regularization and weights' nonnegativity, (iii) drawing more insight into variety of dataset, (iv) comparing the proposed with recent AE architectures, and lastly (v) supporting the interpretability claim with new experiments on text categorization data. The paper is structured as follows: Section II introduces the network configuration and the notation for nonnegative sparse feature extraction. Section III discusses the experimental designs and Section IV presents the results. Finally, conclusions are drawn in Section V.
\section{Nonnegative sparse feature extraction using Constrained Autoencoders}
As shown in \cite{lee1999learning}, one way of representing data is by shattering it into various distinct pieces in a manner that additive merging of these pieces can reconstruct the original data. Mapping this intuition to AEs, the idea is to sparsely disintegrate data into parts in the encoding layer and subsequently additively process the parts to recombine the original data in the decoding layer. This disintegration can be achieved by imposing nonnegativity constraint on the network's weights \cite{wright1999,Nguyen2013Learning,Ehsan2015Deep}.
\begin{figure*}[htb!]
	\begin{minipage}[b]{0.245\linewidth}
		\centering
		\centerline{\includegraphics[scale=0.55]{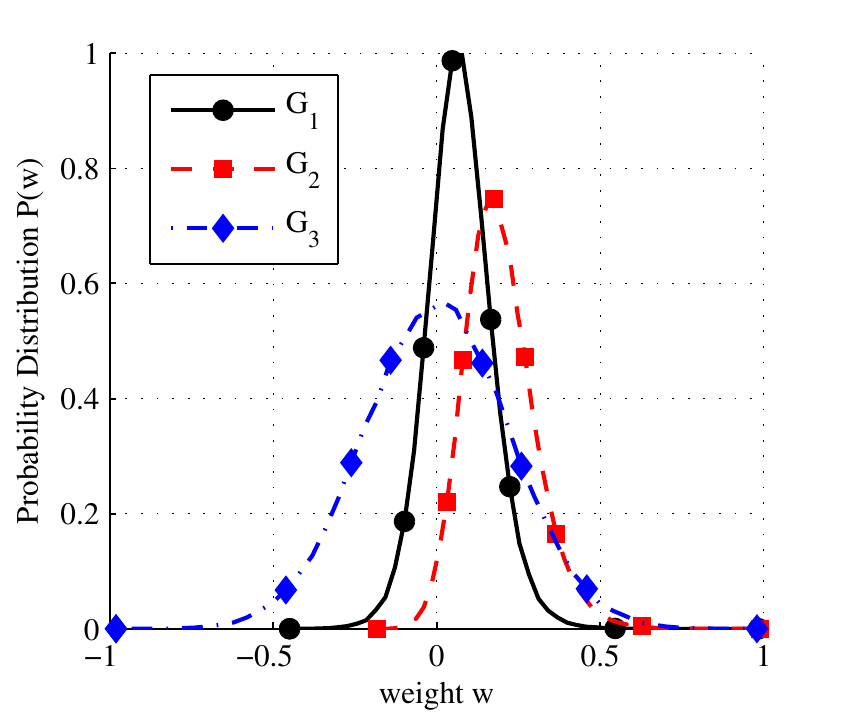}}
		{{\footnotesize (a)}}
	\end{minipage}
\begin{minipage}[b]{0.245\linewidth}
		\centering
		\centerline{\includegraphics[scale=0.55]{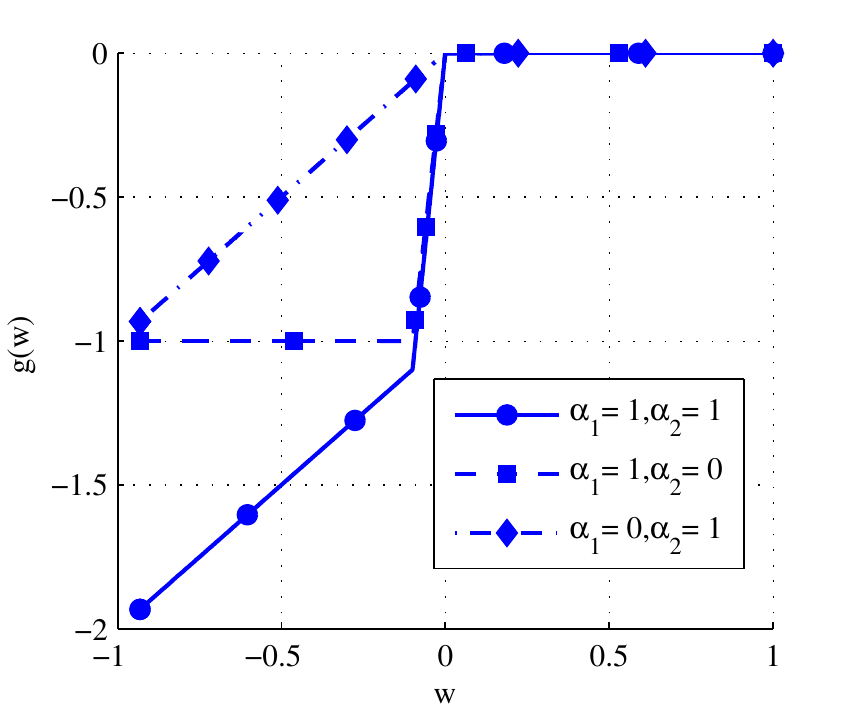}} %
		{{\footnotesize(b)}}
	\end{minipage}
\begin{minipage}[b]{0.245\linewidth}
		\centering
		\centerline{\includegraphics[scale=0.55]{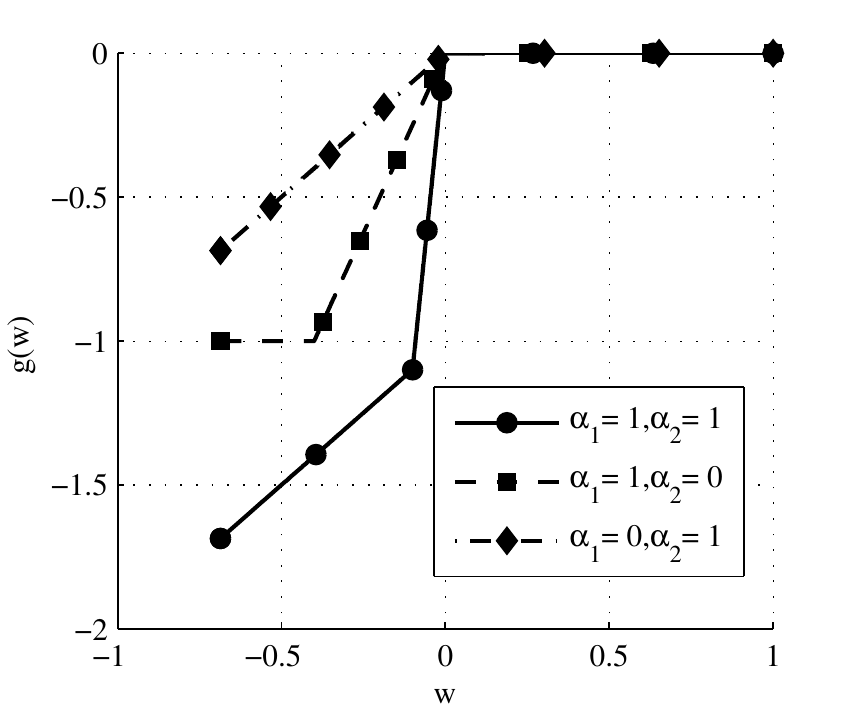}} %
		{{\footnotesize(c)}}
	\end{minipage}
\begin{minipage}[b]{0.245\linewidth}
		\centering
		\centerline{\includegraphics[scale=0.55]{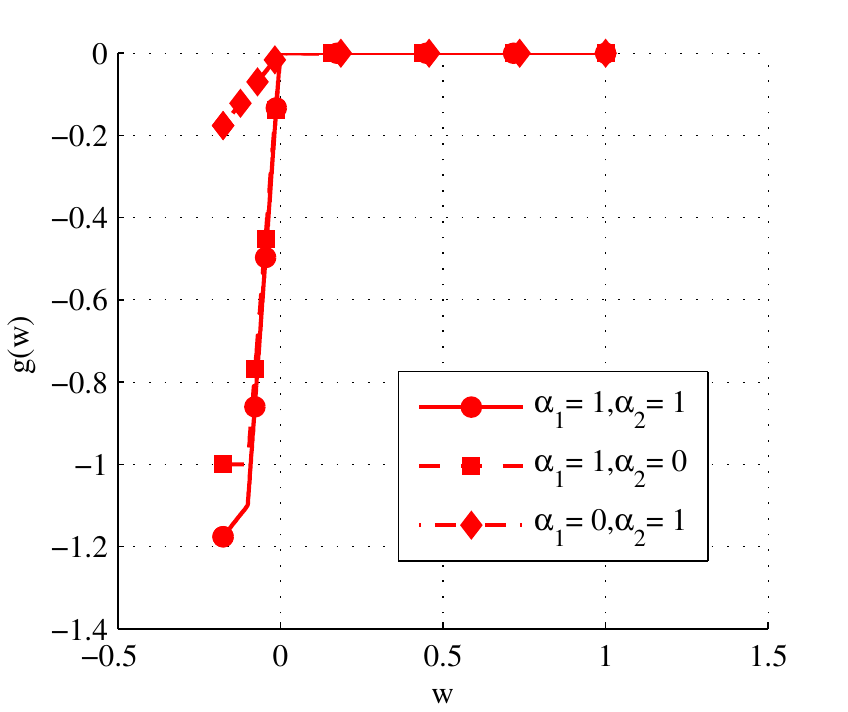}} %
		{{\footnotesize(d)}}
	\end{minipage}
	\caption{(a) Symmetric ($G_3$) and skewed ($G_1$ and $G_2$) weight distributions. Decay function with three values of $\alpha_1$ and $\alpha_2$ for weight distribution (b) $G_3$ (c) $G_1$ and (d) $G_2$.}\label{weight_dist_example}
\end{figure*}
\subsection{$L_1/L_2$-Nonnegativity Constrained Sparse Autoencoder ($L_1/L_2$-NCSAE)}
In order to encourage higher degree of nonnegativity in network's weights, a composite penalty term \eqref{MyEq8} is added to the objective function resulting in the cost function expression for $L_1/L_2$-NCSAE:
\begin{equation}\label{MyEq8}
\begin{split}
J_{\text{$L_1/L_2$-NCSAE}}\big(\textbf{W},\textbf{b}\big) = J_{AE} &+\beta \sum^{n'}_{r=1} D_{KL}\bigg(p\bigg\Vert \frac{1}{m}\sum_{k=1}^mh_r(\textbf{x}^{(k)})\bigg)\\
& +\sum_{l=1}^{2}\sum_{i=1}^{s_{l}}\sum_{j=1}^{s_{l+1}}f_{L_1/L_2}\big(w_{ij}^{(l)}\big) %
\end{split}
\end{equation}
where $\textbf{W} = \{\textbf{W}^{(1)},\textbf{W}^{(2)}\}$ and $\textbf{b} = \{\textbf{b}_x,\textbf{b}_h\}$ represent the weights and biases of encoding and decoding layers respectively; $s_l$ is the number of neurons in layer $l$. $w^{(l)}_{ij}$ represents the connection between $j$th neuron in layer $l-1$ and $i$th neuron in layer $l$ and for given input  $\textbf{x}$,
\begin{equation}\label{MyEq9zz}
J_{AE} = \frac{1}{m}\sum^m_{k=1}\norm{\sigma(\textbf{W}^{(2)} \sigma(\textbf{W}^{(1)}\textbf{x}^{(k)}+\textbf{b}_x)+\textbf{b}_h) - \textbf{x}^{(k)}}^2_2,
\end{equation}
where $m$ is the number of training examples, $||\centerdot||_2$ is the Euclidean norm. $D_{KL}(\centerdot)$ is the Kullback-Leibler (KL) divergence for sparsity control \cite{ng2011ufldl} with $p$ denoting the desired activation and the average activations of hidden units, $n'$ is the number of hidden units, $h_j(\textbf{x}^{(k)})= \sigma(\textbf{W}_j^{(1)}\textbf{x}^{(k)}+b_{x,j})$ denotes the activation of hidden unit $j$ due to input $\textbf{x}^{(k)}$, and $\sigma(\centerdot)$ is the element-wise application of the logistic sigmoid, $\sigma(\textbf{x})=\sfrac{1}{(1+exp(-\textbf{x}))}$, $\beta$ controls the sparsity penalty term, and
\begin{equation}\label{MyEq9}
f_{L_1/L_2}(w_{ij}) = \Bigg\{
\begin{array}{l l}
\alpha_1\Gamma(w_{ij},\kappa) + \frac{\alpha_2}{2}||w_{ij}||^2 & \quad w_{ij} <0\\
0 & \quad w_{ij} \geq 0
\end{array}
\end{equation}
where $\alpha_1$ and $\alpha_2$ are $L_1$ and $L_2$ nonnegativity-constraint weight penalty factors, respectively. $p$, $\beta$, $\alpha_1$, and $\alpha_2$ are experimentally set to $0.05$, $3$, $0.0003$, and $0.003$, respectively using $9000$ randomly sampled images from the training set as a held-out validation set for hyperparameter tuning and the network is retrained on the entire dataset. The weights are updated as below using the error backpropagation:
\begin{equation}\label{MyEq10}
w_{ij}^{(l)} =w_{ij}^{(l)}-\xi \frac{\partial}{\partial w_{ij}^{(l)}}J_{\text{$L_1/L_2$-NCSAE}}(\textbf{W},\textbf{b})
\end{equation}
\begin{equation}\label{MyEq11}
b_{i}^{(l)}=b_{i}^{(l)}-\xi \frac{\partial}{\partial b_{i}^{(l)}}J_{\text{$L_1/L_2$-NCSAE}}(\textbf{W},\textbf{b})
\end{equation}
where $\xi>0$ is the learning rate and the gradient of $L_1/L_2$-NCSAE loss function is computed as in (\ref{MyEq12}).
\begin{equation}\label{MyEq12}
\begin{split}
\frac{\partial}{\partial w_{ij}^{(l)}}J_{\text{$L_1/L_2$-NCSAE}}(\textbf{W},\textbf{b}) &=\frac{\partial}{\partial w_{ij}^{(l)}}J_{\text{AE}}\big(\textbf{W},\textbf{b}\big)\\
&+\beta \frac{\partial}{\partial w_{ij}^{(l)}}D_{KL}\bigg(p\bigg\Vert \frac{1}{m}\sum_{k=1}^mh_j(\textbf{x}^{(k)})\bigg)\\
&+ g\big(w_{ij}^{(l)}\big)
\end{split}
\end{equation}
where $g(w_{ij})$ is a composite function denoting the derivative of $f_{L_1/L_2}(w_{ij})$ \eqref{MyEq9} with respect to $w_{ij}$ as in \eqref{MyEq13}.
\begin{equation}\label{MyEq13}
g(w_{ij}) = \bigg\{
\begin{array}{l l}
\alpha_1\nabla_{\textbf{w}}\norm{w_{ij}} + \alpha_2w_{ij} & \quad w_{ij} <0\\
0 & \quad w_{ij}  \geq 0
\end{array}
\end{equation}\\\\
\indent
Although the penalty function in \eqref{MyEq8} is an extension of NCAE (obtained by setting $\alpha_1$ to zero), a close scrutiny of the weight distribution of both the encoding and decoding layer in NCAE reveals that many weights are still not nonnegative despite imposing nonnegativity constraints. The reason for this is that the original \emph{$L_2$} norm used in NCAE penalizes the negative weights with big magnitudes stronger than those with smaller magnitudes. This forces a good number of the weights to take on small negative values. This paper uses additional \emph{$L_1$} to even out this occurrence, that is, the \emph{$L_1$} penalty forces most of the negative weights to become nonnegative.
\subsection{Implication of imposing nonnegative parameters with composite decay function}
The graphical illustration of the relation between the weight distribution and the composite decay function is shown in Fig.~\ref{weight_dist_example}. Ideally, addition of Frobenius norm of the weight matrix ($\alpha||\textbf{W}||_F^2$) to the reconstruction error in \eqref{MyEq9zz} imposes a Gaussian prior on the weight distribution as shown in curve $G_3$ in Fig.~\ref{weight_dist_example}a. However, using the composite function in \eqref{MyEq9} results in imposition of positively-skewed deformed Gaussian distribution as in curves $G_1$ and $G_2$. The degree of nonnegativity can be adjusted using parameters $\alpha_1$ and $\alpha_2$. Both parameters have to be carefully chosen to enforce nonnegativity while simultaneously ensuring good supervised learning outcomes. The effect of $L_1$ ($\alpha_2=0$), $L_2$ ($\alpha_1=0$) and $L_1/L_2$ ($\alpha_1\neq0$ and $\alpha_2\neq0$) nonnegativity penalty terms on weight updates for weight distributions $G_1$, $G_2$ and $G_3$ are respectively shown in Fig.~\ref{weight_dist_example}c,d, and b. It can be observed for all the three distributions that $L_1/L_2$ regularization enforces stronger weight decay than individual $L_1$ and $L_2$ regularization. Other observation from Fig.~\ref{weight_dist_example} is that the more positively-skewed the weight distribution becomes, the lesser the weight decay function.\\
\indent
The consequences of minimizing \eqref{MyEq8} are that: (i) the average reconstruction error is reduced (ii) the sparsity of the hidden layer activations is increased because more negative weights are forced to zero thereby leading to sparsity enhancement, and (iii) the number of nonnegative weights is also increased. The resultant effect of penalizing the weights simultaneously with \emph{$L_1$} and \emph{$L_2$} norm is that large positive connections are preserved while their magnitudes are shrunk. However, the $L_1$ norm in \eqref{MyEq9} is non-differentiable at the origin, and this can lead to numerical instability during simulations. To circumvent this drawback, one of the well known smoothing function that approximates \emph{$L_1$} norm as in \eqref{MyEq9} is utilized. Given any finite dimensional vector $\textbf{z}$ and positive constant $\kappa$, the following smoothing function approximates \emph{$L_1$} norm:
\begin{equation}\label{MyEq9d}
\begin{split}
\Gamma(\textbf{z},\kappa) = \Bigg\{
\begin{array}{l l}
||\textbf{z}|| & \quad ||\textbf{z}|| > \kappa \\ \\
\frac{||\textbf{z}||^2}{2\kappa} + \frac{\kappa}{2} & \quad ||\textbf{z}|| \leq \kappa
\end{array}
\end{split}
\end{equation}
with gradient
\begin{equation}\label{MyEq9g}
\nabla_{\textbf{z}}\Gamma(\textbf{z},\kappa) = \Bigg\{
\begin{array}{l l}
\frac{\textbf{z}}{||\textbf{z}||} & \quad ||\textbf{z}|| > \kappa \\ \\
\frac{\textbf{z}}{\kappa}  & \quad ||\textbf{z}|| \leq \kappa
\end{array}
\end{equation}
For convenience, we adopt \eqref{MyEq9d} to smoothen the $L_1$ penalty function and $\kappa$ is experimentally set to $0.1$.
\section{Experiments}
\begin{figure*}[htbp!]
  \centering
  \includegraphics[width=0.85\linewidth]{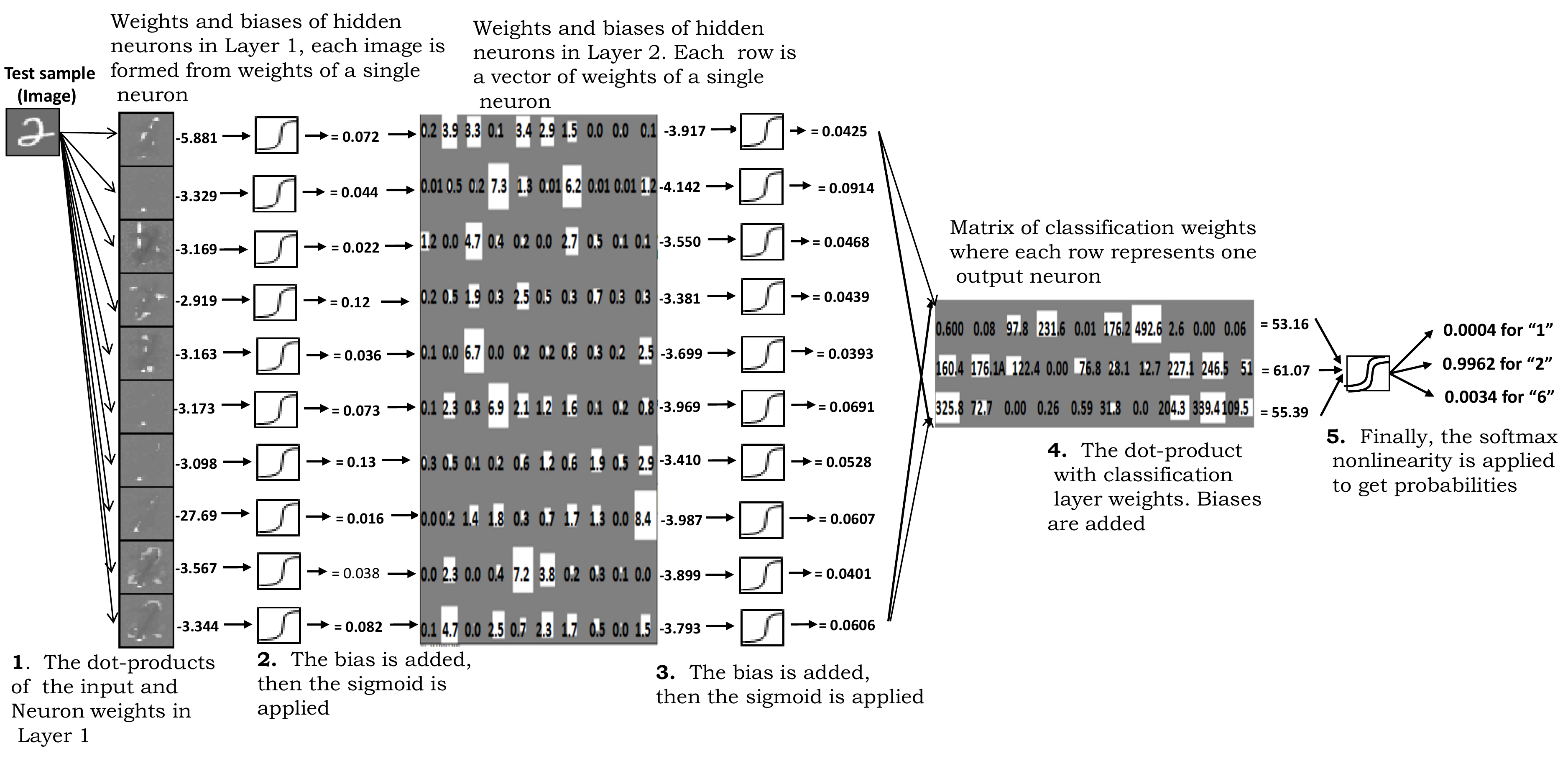}
  \caption{Filtering the signal through the $L_1/L_2$-NCSAE trained using the reduced MNIST data set with class labels $1$, $2$ and $6$. The test image is a 28$\times$28 pixels image unrolled into a vector of 784  values. Both the input test sample and the receptive fields of the first autoencoding layer are presented as images. The weights of the output layer are plotted as a diagram  with one row for each output neuron and one column for every hidden neuron in $(L-1)^{th}$ layer. The architecture is 784-10-10-3. The range of weights are scaled to [-1,1] and mapped to the graycolor map. $w=-1$ is assigned to black, $w=0$ to grey, and $w=1$ is assigned to white color. That is, black pixels indicate negative, grey pixels indicate zero-valued weights and white pixels indicate positive weights.}\label{deep_receptive}
\end{figure*}
In the experiments, three data sets are used, namely: MNIST \cite{LeCun1998}, NORB normalized-uniform \cite{lecun2004learning}, and Reuters-21578 text categorization dataset. The Reuters-21578 text categorization dataset comprises of documents that featured in 1987 Reuters newswire. The ModApte split was employed to limit the dataset to 10 most frequent classes. The ModApte split was utilized to limit the categories to 10 most frequent categories. The bag-of-words format that has been stemmed and stop-word removed was used; see http://people.kyb.tuebingen.mpg.de/pgehler/rap/ for further clarification. The dataset contains $11,413$ documents with $12,317$ dimensions. Two techniques were used to reduce the dimensionality of each document in order to preserve the most informative and less correlated words \cite{tan2006introduction}. To reduce the dimensionality of each document to contain the most informative and less correlated words, words were first sorted based on their frequency of occurrence in the dataset. Words with frequency below 4 and above $70$ were then eliminated. The most informative words that do not occur in every topic were selected based on information gain with the class attribute. The remaining words (or features) in the dataset were sorted using this method, and the less important features were removed based on the desired dimension of documents. In this paper, the length of the feature vector for each of the documents was reduced to 200.\\
\indent
In the preliminary experiment, the subset $1$, $2$ and $6$ from the MNIST handwritten digits as extracted for the purpose of understanding how the deep network constructed using $L_1/L_2$-NCSAE processes and classifies its input. For easy interpretation, a small deep network was constructed and trained by stacking two AEs with $10$ hidden neurons each and $3$ softmax neurons. The number of hidden neurons was chosen to obtain reasonably good classification accuracy while keeping the network reasonably small. The network is intentionally kept small because the full MNIST data would require larger hidden layer size and this may limit network interpretability. An image of digit $2$ is then filtered through the network, and it can be observed in Fig.~\ref{deep_receptive} that sparsification of the weights in all the layers is one of the aftermath of nonnegativity constraints imposed on the network. Another observation is that most of the weights in the network have been confined to nonnegative domain, which removes opaqueness of the deep learning process. It can be seen that the fourth and seventh receptive fields of the first AE layer have dominant activations (with activation values $0.12$ and $0.13$ respectively) and they capture most information about the test input. Also, they are able to filter distinct part of input digit. The outputs of the first layer sigmoid constitute higher level features extracted from test image with emphasis on the fourth and seventh features. Subsequently in second layer the second, sixth, eight, and tenth neurons have dominant activations (with activation values $0.0914$, $0.0691$, $0.0607$, and $0.0606$ respectively) because they have stronger connections with the dominant neurons in first layer than the rest. Lastly in the softmax layer, the second neuron was $99.62\%$ activated because it has strongest connections with the dominant neurons in second layer thereby classifying the test image as "2".\\
\indent
\begin{figure*}[htb!]
	\begin{minipage}[b]{1.0\linewidth}
		\centering
		\centerline{\includegraphics[height=8cm,width=0.85\linewidth]{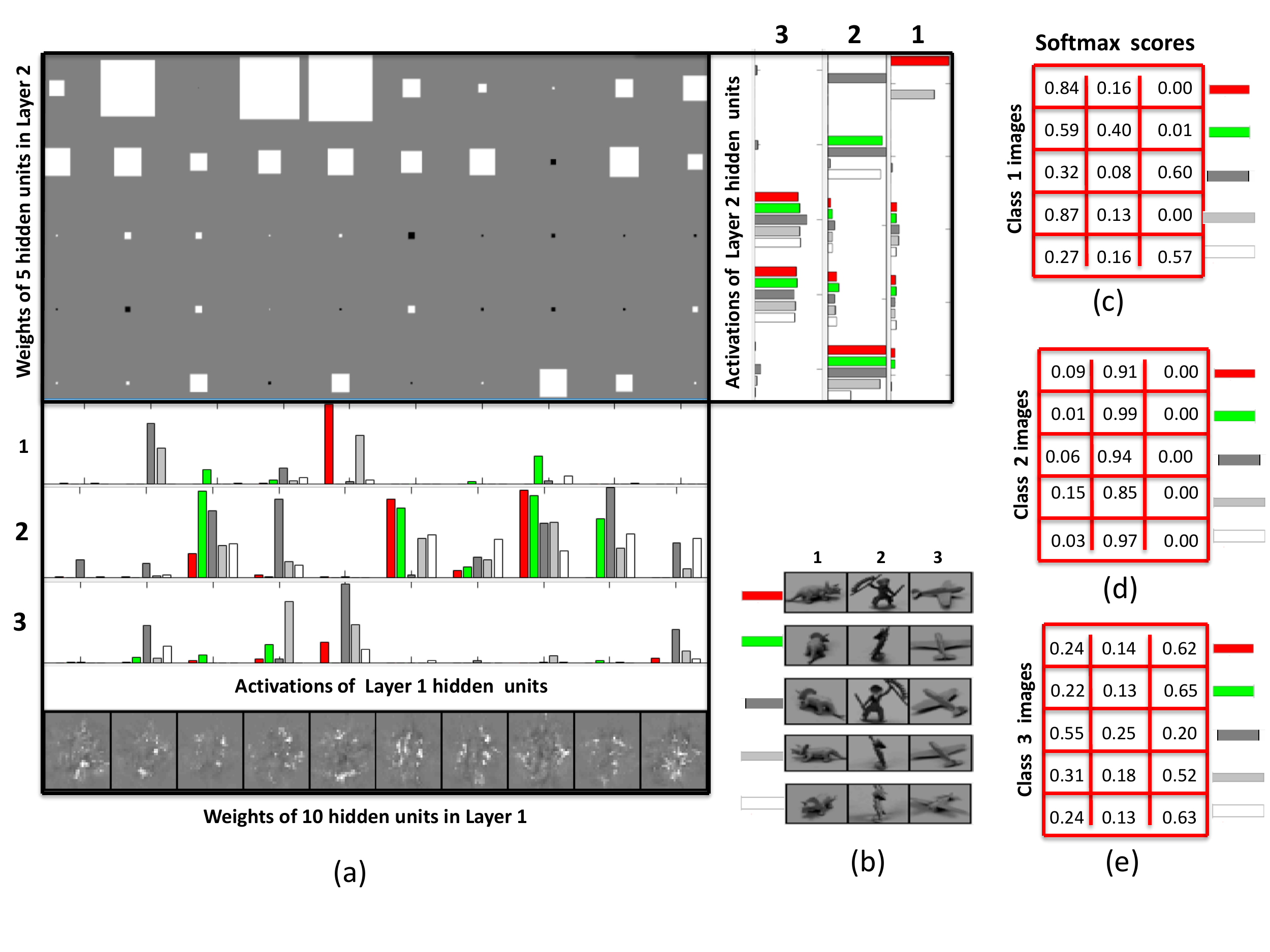}}
	\end{minipage}
	\caption{The weights were trained using two stacked $L_1/L_2$-NCSAEs. RFs learned from the reduced NORB dataset are plotted as images at the bottom part of (a). The intensity of each pixel is proportional to the magnitude of the weight connected to that pixel in the input image with negative value indicating black, positive values white, and the value 0 corresponding to gray. The biases are not shown. The activations of first layer hidden units for the NORB objects presented in (b) are depicted on the bar chart on top of the RFs. The weights of the second layer AE are plotted as a diagram at the topmost part of (a). Each row of the plot corresponds to the weight of each hidden unit of second AE and each column for weight of every hidden unit of the first layer AE. The magnitude of the weight corresponds to the area of each square; white indicates positive, grey indicates zero, and black negative sign. The activations of second layer hidden units are shown as bar chart in the right-hand side of the second layer weight diagram. Each column shows the activations of each hidden unit for five color-coded examples of the same object. The outputs of Softmax layer for color-coded test objects with class labels (c) "fourlegged animals" tagged as class 1, (d) "human figures" as class 2, and (e) "airplanes" as class 3. }\label{norb_mag}		
\end{figure*}
The fostering of interpretability is also demonstrated using a subset of NORB normalized-uniform dataset \cite{lecun2004learning} with class labels "four-legged animals", "human figures", "airplanes". The $1024$-$10$-$5$-$3$ network configuration was trained on the subset of the NORB data using two stacked $L_1/L_2$-NCSAEs and a Softmax layer. Fig.~\ref{norb_mag}b shows the randomly sampled test patterns and the weights and activations of first and second AE layer are shown in Fig.~\ref{norb_mag}a. The bar charts indicate the activations of hidden units for the sample input patterns. The features learned by units in each layer are localized, sparse and allow easy interpretation of isolated data parts. The features mostly show nonnegative weights making it easier to visualize to what input object patterns they respond. It can be seen that units in the network discriminate among objects in the images and react differently to input patterns. Third, sixth, eight, and ninth hidden units of layer 1 capture features that are common to objects in class "2" and react mainly to them as shown in the first layer activations. Also, the features captured by the second layer activations reveal that second and fifth hidden units are mainly stimulated by objects in class "2".\\
\indent
The outputs of Softmax layer represent the \emph{a posteriori} class probabilities for a given sample and are denoted as Softmax scores. An important observation from Fig.~\ref{norb_mag}a,b, and c is that hidden units in both layers did not capture significant representative features for class "1" white color-coded test sample. This is one of the reasons why it is misclassified into class "3" with probability of 0.57. The argument also goes for class "1" dark-grey color-coded test sample misclassified into class "3" with probability of 0.60. In contrast, hidden units in both layers capture significant representative features for class "2" test samples of all color codes. This is why all class "2" test samples are classified correctly with high probabilities as shown in Fig.~\ref{norb_mag}d. Lastly, the network contains a good number of representative features for class "3" test samples and was able to classify 4 out of 5 correctly as given in Fig.~\ref{norb_mag}e.

\section{Results and Discussion}
\subsection{Unsupervised Feature Learning of Image Data}
\begin{figure*}[htb!]	
	\begin{minipage}[b]{1.0\linewidth}
		\centering
		\centerline{\includegraphics[width=0.85\linewidth]{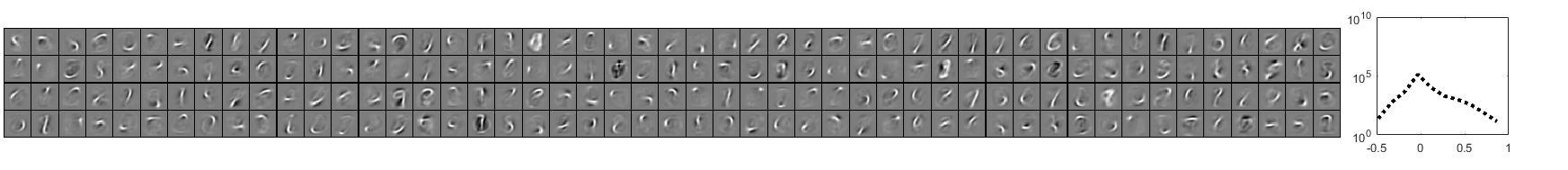}}
		\vspace{-5mm}
		{{\footnotesize (a) SAE}}
	\end{minipage}
	\begin{minipage}[b]{1.0\linewidth}
		\centering
		\vspace{1mm}
		\centerline{\includegraphics[width=0.85\linewidth]{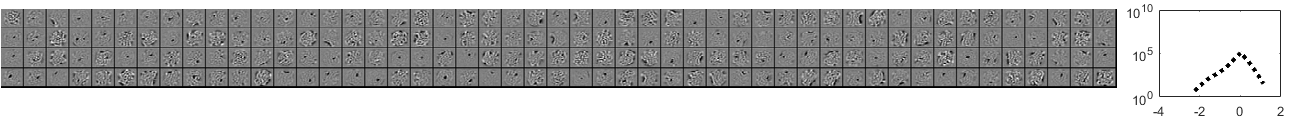}}
		\vspace{-5mm}
		{{\footnotesize(b) DpAE}}
	\end{minipage}
\begin{minipage}[b]{1.0\linewidth}
		\centering
		\vspace{1mm}
		\centerline{\includegraphics[width=0.85\linewidth]{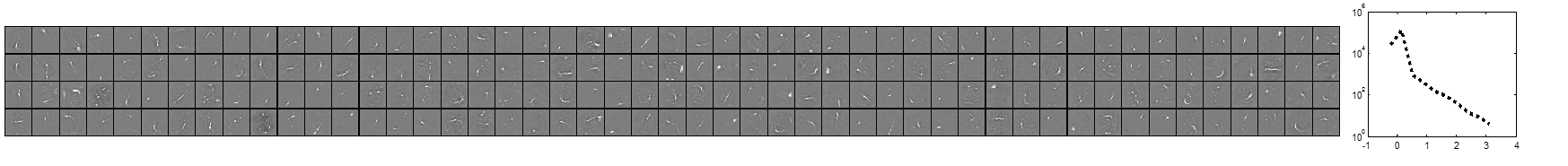}}
		\vspace{-5mm}
		{{\footnotesize(c) NCAE}}
	\end{minipage}
\begin{minipage}[b]{1.0\linewidth}
		\centering
		\vspace{1mm}
		\centerline{\includegraphics[width=0.85\linewidth]{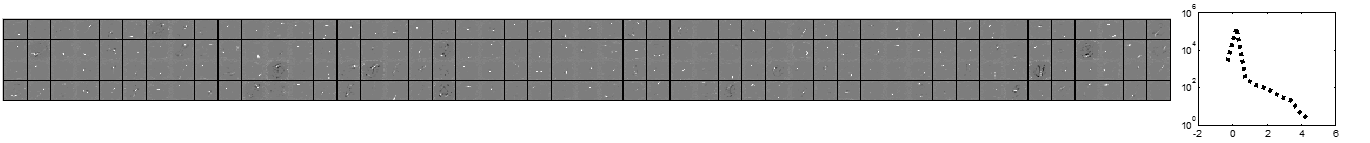}}
		\vspace{-5mm}
		{{\footnotesize(d) $L_1/L_2$-NCSAE}}
	\end{minipage}
	\caption{196 receptive fields ($\mathbf{W}^{(1)}$) with weight histograms learned from MNIST digit data set using (a) SAE, (b)  DpAE  (c) NCAE, and (d) $L_1/L_2$-NCSAE. Black pixels indicate negative, and white pixels indicate positive weights. The range of weights are scaled to [-1,1] and mapped to the graycolor map. $w=-1$ is assigned to black, $w=0$ to grey, and $w=1$ is assigned to white color.}
	\label{Receptive_fields_MNIST}
\end{figure*}

In the first set of experiments, three-layer $L_1/L_2$-NCSAE, NCAE \cite{Ehsan2015Deep}, DpAE \cite{hinton2012improving}, and conventional SAE network with $196$ hidden neurons were trained using MNIST dataset of handwritten digits and their ability to discover patterns in high dimensional data are compared. These experiments were run one time and recorded. The encoding weights $\textbf{W}^{(1)}$, also known as receptive fields or filters as in the case of image data, are reshaped, scaled, centered in a 28 $\times$ 28 pixel box and visualized. The filters learned by $L_1/L_2$-NCSAE are compared with that learned by its counterparts, NCAE and SAE. It can be easily observed from the results in Fig.~\ref{Receptive_fields_MNIST} that $L_1/L_2$-NCSAE learned receptive fields that are more sparse and localized than those of SAE, DpAE, and NCAE. It is remarked that the black pixels in both SAE and DpAE features are results of the negative weights whose values and numbers are reduced in NCAE with nonnegativity constraints, which are further reduced by imposing an additional $L_1$ penalty term in $L_1/L_2$-NCSAE as shown in the histograms located on the right side of the figure. In the case of $L_1/L_2$-NCSAE, tiny strokes and dots which constitute the basic part of handwritten digits, are unearthed compared to SAE, DpAE, and NCAE. Most of the features learned by SAE are major parts of the digits or the blurred version of the digits, which are obviously not as sparse as those learned by $L_1/L_2$-NCSAE. Also, the features learned by DpAE are fuzzy compared to those of $L_1/L_2$-NCSAE which are sparse and distinct. Therefore, the achieved sparsity in the encoding can be traced to the ability of $L_1$ and $L_2$ regularization in enforcing high degree of weights' nonnegativity in the network.\\
\begin{figure*}[htb!]
		\begin{minipage}[b]{0.5\linewidth}
		\centering
       \captionsetup{justification=centering}
		\centerline{\includegraphics[scale=0.5]{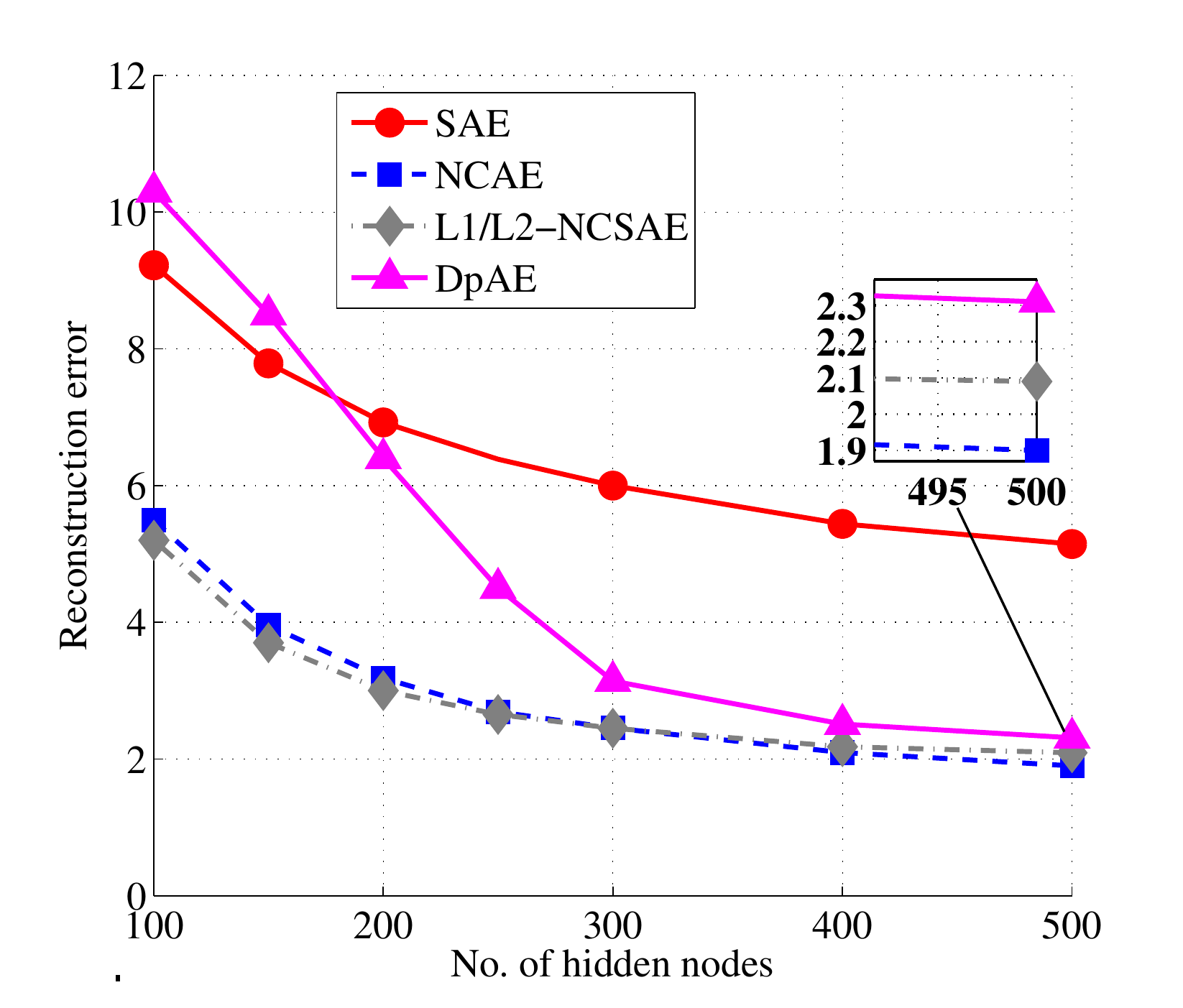}}
		{{\footnotesize (a)}}
	\end{minipage}
	\begin{minipage}[b]{0.5\linewidth}
		\centering
		\centerline{\includegraphics[scale=0.5]{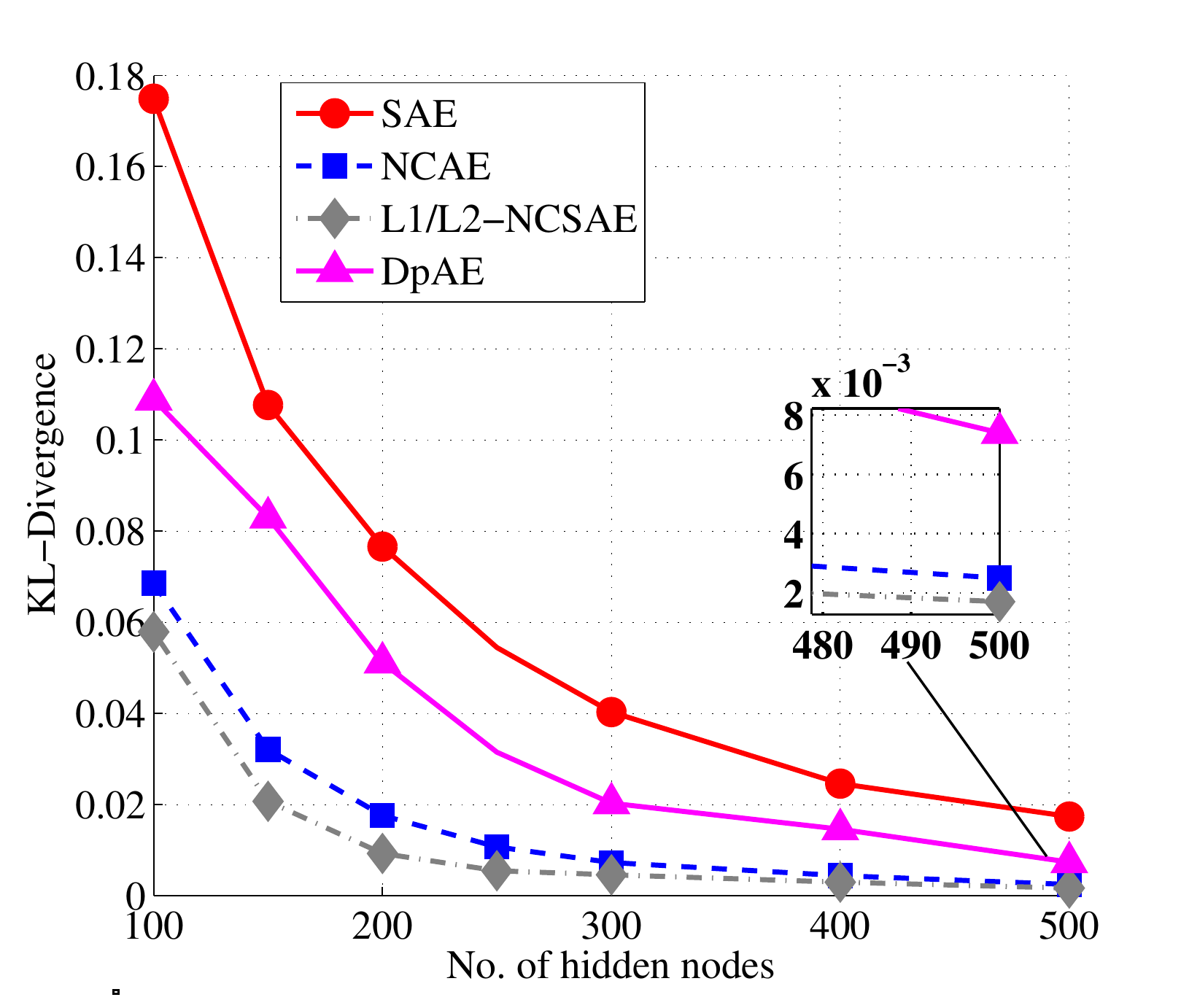}}
		{{\footnotesize(b)}}
	\end{minipage}
	\caption{(a) Reconstruction error and (b) Sparsity of hidden units measured by KL-divergence using MNIST train dataset with $p$ = 0.05. }\label{hidden_size}		
\end{figure*}
\begin{figure*}[htb!]
  \centering
  \includegraphics[width=1.0\linewidth]{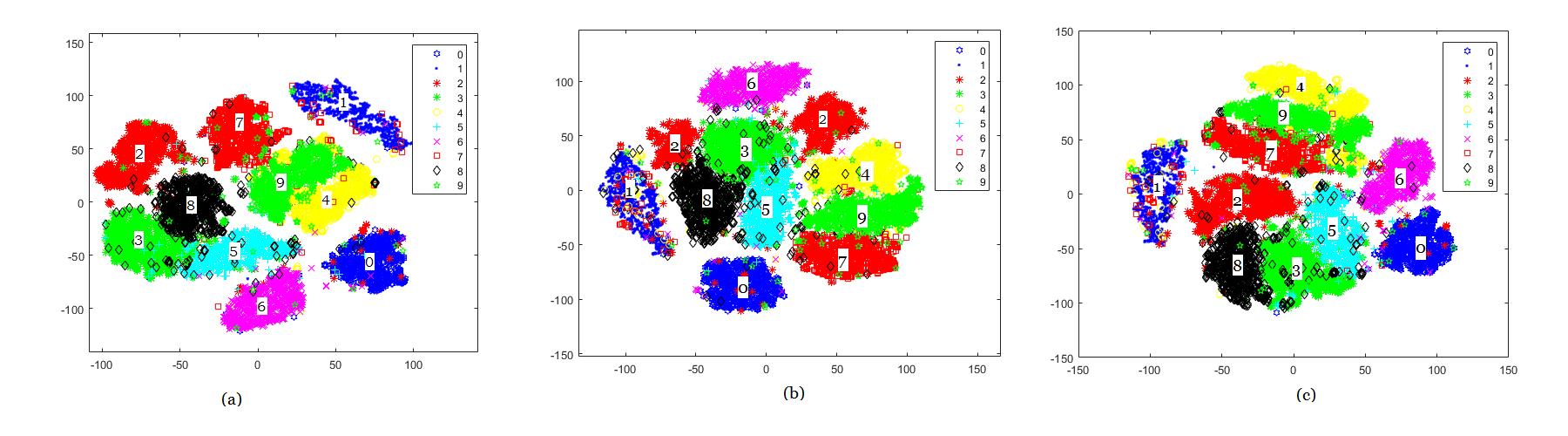}
  \caption{t-SNE projection \cite{Maaten2008} of 196D representations of MNIST handwritten digits using (a) DpAE (b) NCAE (c) $L_1/L_2$-NCSAE.}\label{tsne_mnist_196_composite}
\end{figure*}
\begin{figure*}[htb!]	
	\begin{minipage}[b]{1.0\linewidth}
		\centering
		\centerline{\includegraphics[width=0.85\linewidth,height=2.0cm]{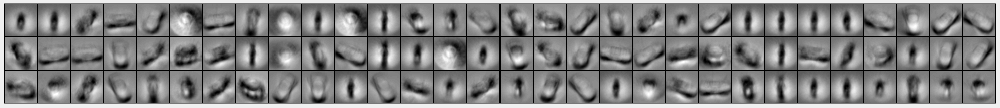}}
		\vspace{-2mm}
		{{\footnotesize (a) SAE}}
	\end{minipage}	
	\begin{minipage}[b]{1.0\linewidth}
		\centering
		\vspace{1mm}
		\centerline{\includegraphics[width=0.85\linewidth,height=2.0cm]{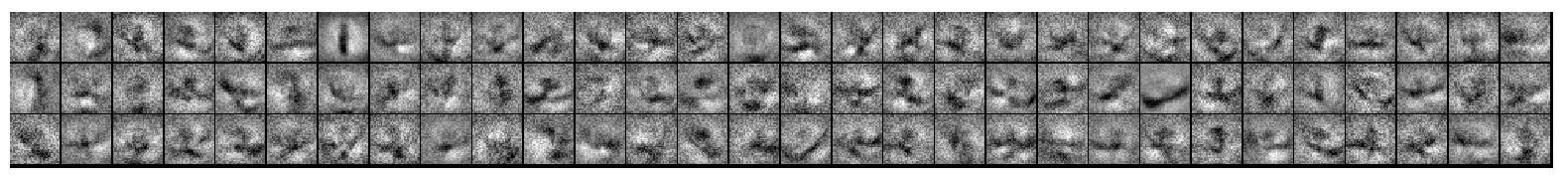}}
		\vspace{-2mm}
		{{\footnotesize(b) DpAE}}
	\end{minipage}
	\begin{minipage}[b]{1.0\linewidth}
		\centering
		\vspace{1mm}
		\centerline{\includegraphics[width=0.85\linewidth,height=2.0cm]{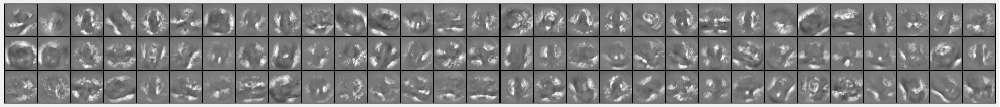}}
		\vspace{-2mm}
		{{\footnotesize(c) NCAE}}
	\end{minipage}
    \begin{minipage}[b]{1.0\linewidth}
		\centering
		\vspace{1mm}
		\centerline{\includegraphics[width=0.85\linewidth,height=2.0cm]{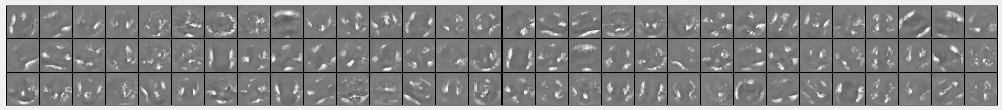}}
		\vspace{-2mm}
		{{\footnotesize(d) $L_1/L_2$-NCSAE}}
	\end{minipage}
	\caption{Weights of randomly selected 90 out of 200 receptive filters of (a) SAE (b) DpAE (c) NCAE, and (d) $L_1/L_2$-NCSAE using NORB dataset. The range of weights are scaled to [-1,1] and mapped to the graycolor map. $w<=-1$ is assigned to black, $w=0$ to grey, and $w>=1$ is assigned to white color.}
	\label{Receptive_fields_NORB}
\end{figure*}
\begin{figure*}[htb!]
	\begin{minipage}[b]{0.33\linewidth}
		\centering
		\centerline{\includegraphics[scale=0.35,height=3.5cm]{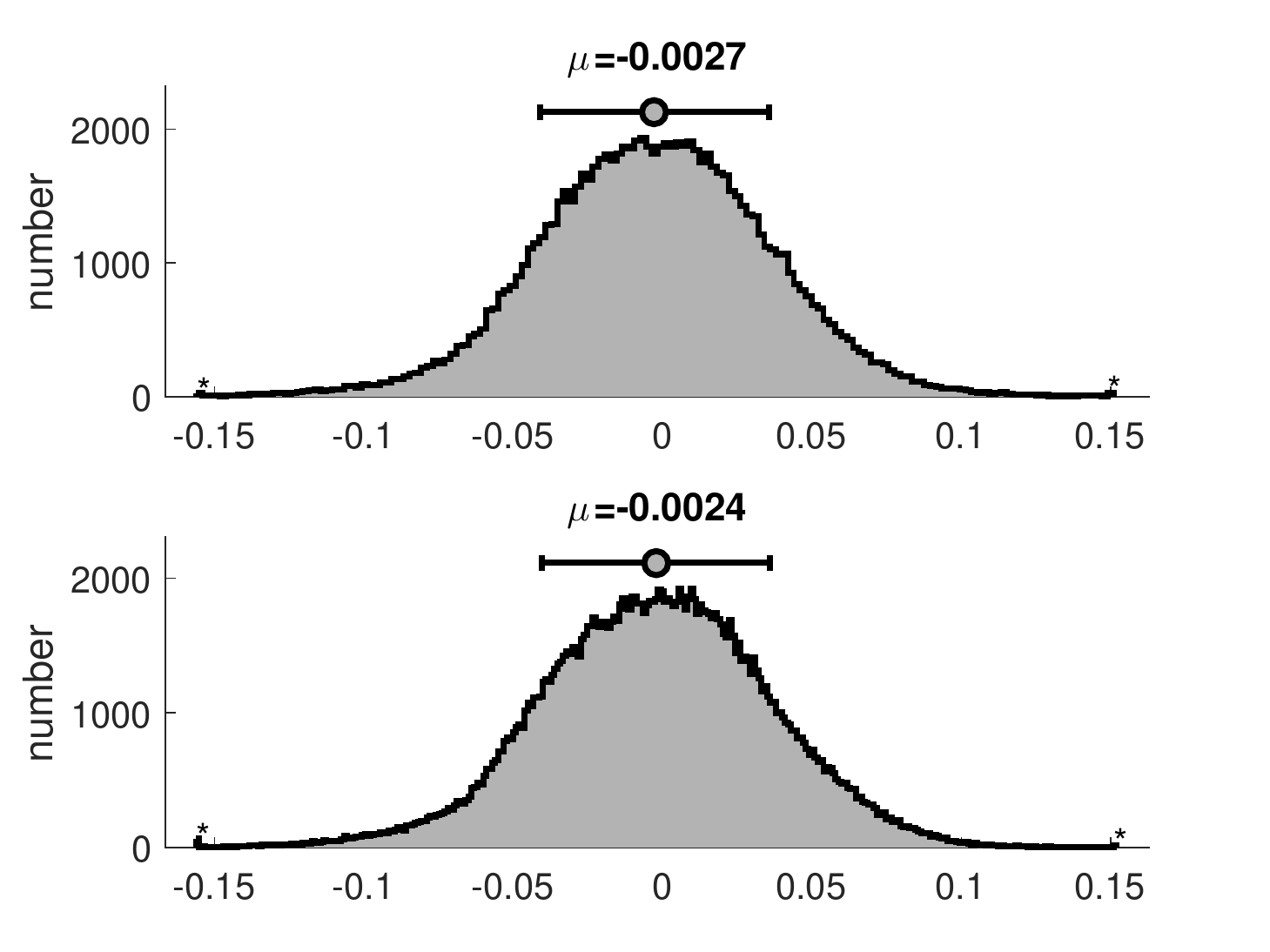}}
		{{\footnotesize (a)}}
	\end{minipage}
\begin{minipage}[b]{0.3\linewidth}
		\centering
		\centerline{\includegraphics[scale=0.35,height=3.5cm]{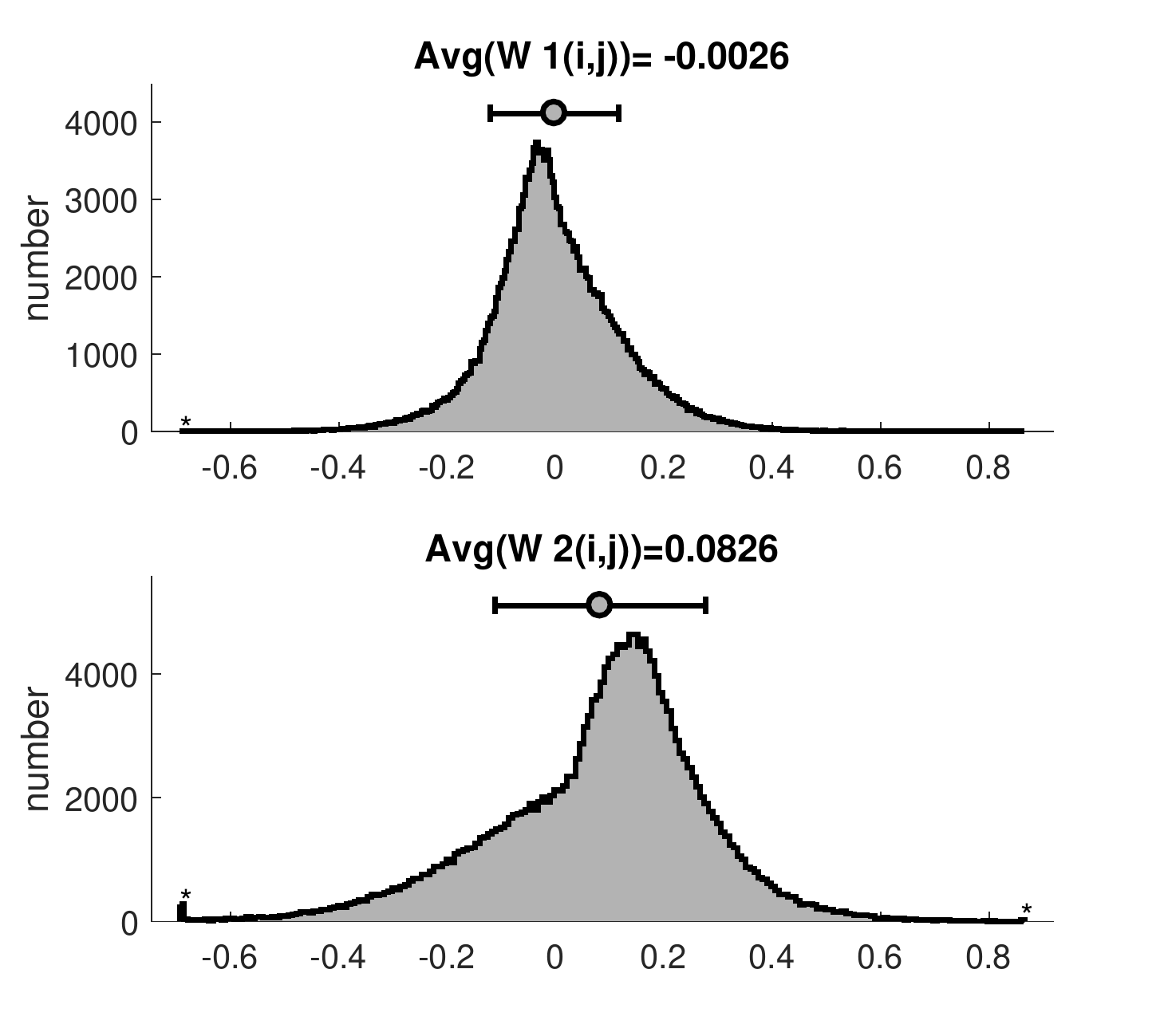}} %
		{{\footnotesize(b)}}
	\end{minipage}
\begin{minipage}[b]{0.33\linewidth}
		\centering
		\centerline{\includegraphics[scale=0.35,height=3.5cm]{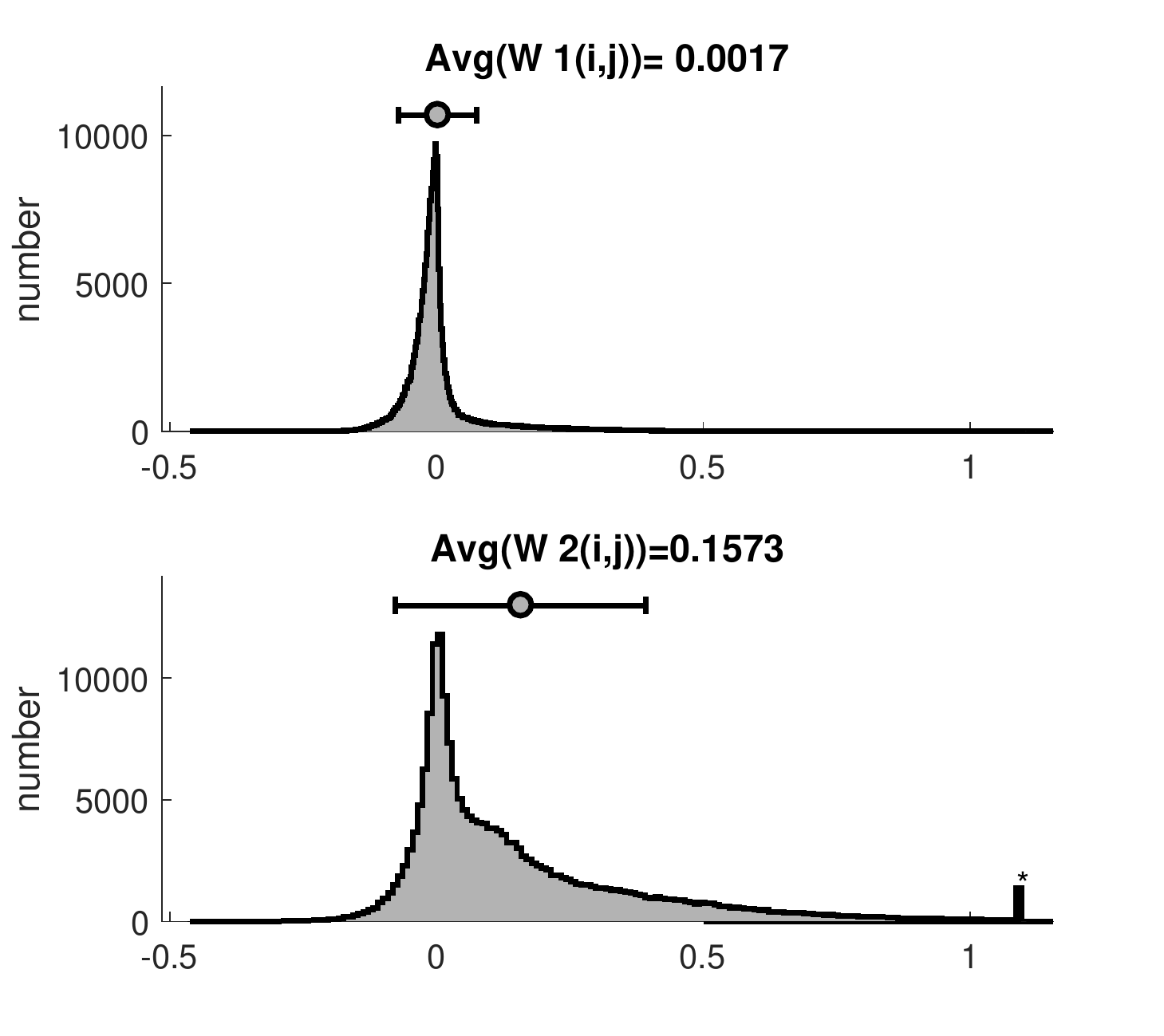}} %
		{{\footnotesize(c)}}
	\end{minipage}
	\caption{ The distribution of 200 encoding ($\mathbf{W}^{(1)}$) and decoding filters ($\mathbf{W}^{(2)}$) weights learned from NORB dataset using (a) DpAE  (b) NCAE (c) $L_1/L_2$-NCSAE. }\label{weight_distribution_norb}
\end{figure*}
\begin{figure*}[htb!]
  \centering
  \includegraphics[width=0.85\textwidth]{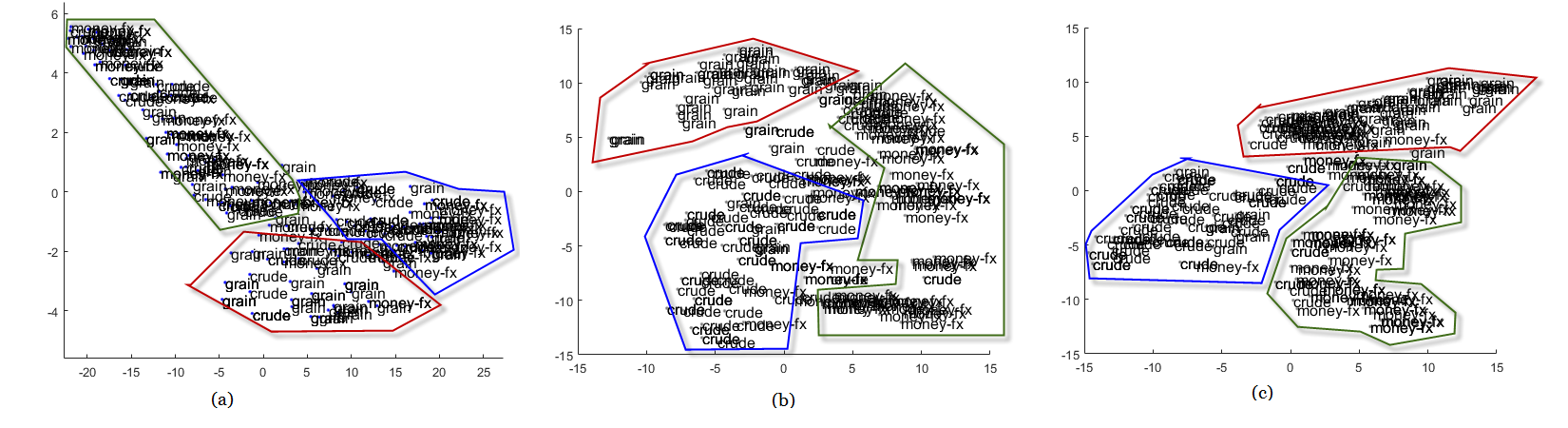}
  \caption{Visualizing 20D representations of a subset of Reuters Documents data using (a) DpAE, (b) NCAE, and (c) $L_1/L_2$-NCSAE. }\label{tsne_reuters_15_compo1}
\end{figure*}
\indent
Likewise in Fig.~\ref{hidden_size}a, $L_1/L_2$-NCSAE with other AEs are compared in terms of reconstruction error, while varying the number of hidden nodes. As expected, it can be observed that $L_1/L_2$-NCSAE yields a reasonably lower reconstruction error on the MNIST training set compared to SAE, DpAE, and NCAE. Although, a close scrutiny of the result also reveals that the reconstruction error of $L_1/L_2$-NCSAE deteriorates compared to NCAE when the hidden size grows beyond $400$. However on the average, $L_1/L_2$-NCSAE reconstructs better than other AEs considered. It can also be observed that DpAE with 50\% dropout has high reconstruction error when the hidden layer size is relatively small (100 or less). This is because the few neurons left are unable to capture the dynamics in the data, which subsequently results in underfitting the data. However, the reconstruction error improves as the hidden layer size is increased. Lower reconstruction error in the case of $L_1/L_2$-NCSAE and NCAE is an indication that nonnegativity constraint facilitates the learning of parts of digits that are essential for reconstructing the digits. In addition, the KL-divergence sparsity measure reveals that $L_1/L_2$-NCSAE has more sparse hidden activations than SAE, DpAE and NCAE for different hidden layer size as shown in Fig.~\ref{hidden_size}b. Again, averaging over all the training examples, $L_1/L_2$-NCSAE yields less activated hidden neurons compared to its counterparts.
\begin{figure*}[htb!]
\begin{minipage}[b]{0.5\linewidth}
		\centering
		\centerline{\includegraphics[scale=0.5,height = 10cm]{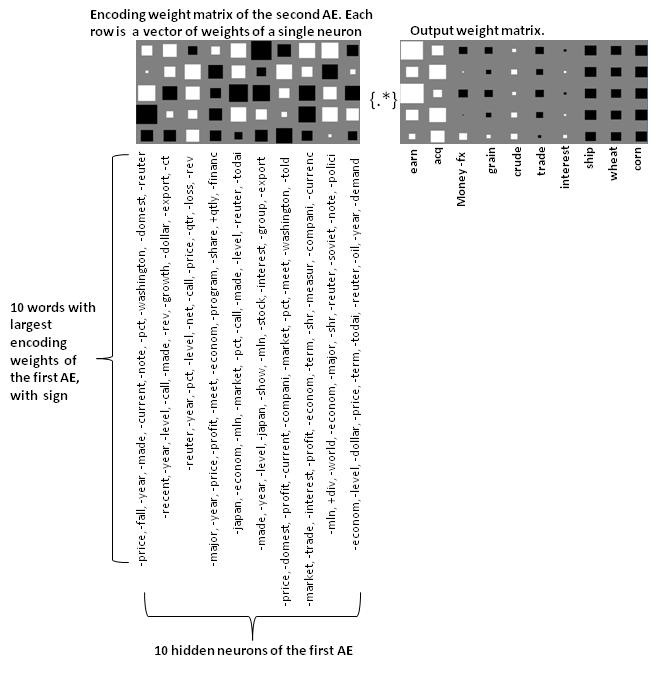}}
		{{\footnotesize (a)}}
	\end{minipage}
	\begin{minipage}[b]{0.5\linewidth}
		\centering
		\centerline{\includegraphics[scale=0.5,height = 10cm]{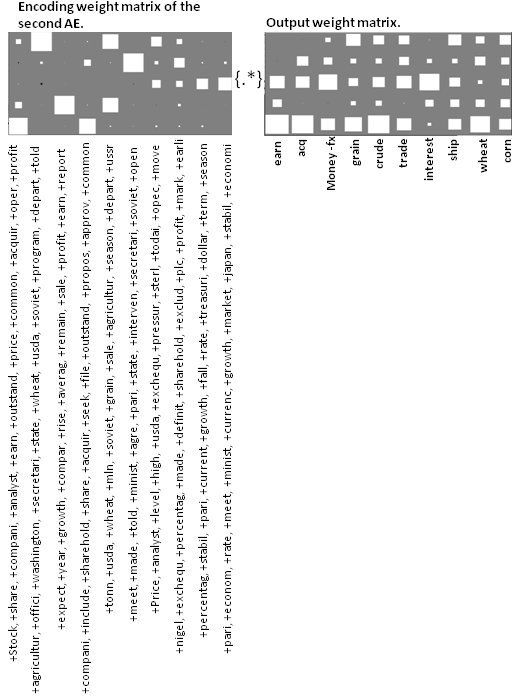}}
		{{\footnotesize (b)}}
	\end{minipage}
	\caption{Deep network trained on Reuters-21578 data using (a) DpAE, (b) $L_1/L_2$-NCSAE. The area of each square is proportional to the weight's magnitude. The range of weights are scaled to [-1,1] and mapped to the graycolor map. $w=-1$ is assigned to black, $w=0$ to grey, and $w=1$ is assigned to white color.}\label{reuters}
\end{figure*}
Also, using t-distributed stochastic neighbor embedding (t-SNE) to project the $196$-D representation of MNIST handwritten digits to 2D space, the distribution of features encoded by $196$ encoding filters of DpAE, NCAE, and $L_1/L_2$-NCSAE are respectively visualized in Figs.~\ref{tsne_mnist_196_composite}a, b, and c. A careful look at Fig.~\ref{tsne_mnist_196_composite}a reveals that digits "$4$" and "$9$" are overlapping in DpAE, and this will inevitably increase the chance of misclassifying these two digits. It can also be observed in Fig.~\ref{tsne_mnist_196_composite}b corresponding to NCAE that digit "$2$" is projected with two different landmarks. In sum, the manifolds of digits with $L_1/L_2$-NCSAE are more separable than its counterpart as shown in Fig.~\ref{tsne_mnist_196_composite}c, aiding the classifier to map out the separating boundaries among the digits more easily.\\
\indent
In the second experiment, SAE, NCAE, $L_1/L_2$-NCSAE, and DpAE  with 200 hidden nodes were trained using the NORB normalized-uniform dataset. The NORB normalized-uniform dataset, which is the second dataset, contains $24,300$ training images and $24,300$ test images of $50$ toys from $5$ generic categories: four-legged animals, human figures, airplanes, trucks, and cars. The training and testing sets consist of $5$ instances of each category. Each image consists of two channels, each of size $96\times 96$ pixels. The inner $64\times 64$ pixels of one of the channels cropped out and resized using bicubic interpolation to $32\times 32$ pixels that form a vector with $1024$ entries as the input. Randomly selected weights of $90$ out of $200$ neurons are plotted in Fig.~\ref{Receptive_fields_NORB}. It can be seen that $L_1/L_2$-NCSAE learned more sparse features compared to features learned by all the other AEs considered. The receptive fields learned by $L_1/L_2$-NCSAE captured the real actual edges of the toys while the edges captured by NCAE are fuzzy, and those learned by DpAE and SAE are holistic. As shown in the weight distribution depicted in Fig.~\ref{weight_distribution_norb}, $L_1/L_2$-NCSAE has both its encoding and decoding weights centered around zero with most of its weights positive when compared with those of DpAE and NCAE that have weights distributed almost even on both sides of the origin.
\begin{table*}[htb!]
 \setlength{\tabcolsep}{4pt}
\caption{Classification accuracy on MNIST and NORB dataset}
\scalebox{1.2}{
\centering{
\begin{tabular}{|c |c|c c|c c c|}
\hline\hline
\multicolumn{2}{|c|}{} & \multicolumn{2}{c|}{\textit{Before} fine-tuning} & \multicolumn{3}{c|}{\textit{After} fine-tuning}\\ [0.5ex]
\cline{3-7}
\multicolumn{1}{|c} {Dataset} & & Mean ($\pm$ SD) & \textit{p}-value & Mean ($\pm$ SD) & \textit{p}-value & \# Epochs \\
\hline        	   	   				
\multirow{4}{*}{MNIST} & SAE  & 0.735 $\pm$ 0.015 & \textless 0.001 & \textbf{0.977} $\pm$ 0.0007 & \textless 0.001  & 400 \\[.5ex] 		
\cline{2-7}
& NCAE  & 0.844 ($\pm$0.0085) & 0.0018 & 0.974 ($\pm$0.0012) & 0.812 & 126 \\[.5ex]
\cline{2-7}
& NNSAE  &  0.702 ($\pm$0.027) & \textless 0.0001 & 0.970 ($\pm$0.001) & \textless 0.0001 & 400 \\[.5ex]
\cline{2-7}
& $L_1/L_2$-NCSAE & \textbf{0.847} ($\pm$0.0077) & \text{-} & 0.974 ($\pm$0.0087) & \text{-} & 84 \\[.5ex]
\cline{2-7}
& DAE (50\% input dropout) & 0.551 ($\pm$0.011) & \textless 0.0001 & 0.972 ($\pm$0.0021) &  0.034 & 400 \\[.5ex]
\cline{2-7}
& DpAE (50\% hidden dropout)  & 0.172 ($\pm$0.0021) & \textless 0.0001 & 0.964 ($\pm$0.0017) & \textless 0.0001 & 400 \\[.5ex]
\cline{2-7}
& AAE  & \textit{-} & \textit{-} & 0.912 ($\pm$0.0016) & \textless 0.0001 & 1000
\\\cline{1-7}
\hline\hline
\multirow{4}{*}{NORB} & SAE  & 0.562 $\pm$ 0.0245 & \textless 0.0001 & 0.814 $\pm$ 0.0099 & 0.041 & 400 \\[.5ex] 
\cline{2-7}
& NCAE  & \textbf{0.696} ($\pm$0.021) & 0.406 & \textbf{0.817} ($\pm$0.0095)  & 0.001 & 305 \\[.5ex]
\cline{2-7}
& NNSAE &  0.208 ($\pm$0.025) & \textless 0.0001 &  0.738 ($\pm$ 0.012)  & \textless 0.001 & 400 \\[.5ex]
\cline{2-7}
& $L_1/L_2$-NCSAE  & 0.695 ($\pm$0.0084) & \text{-} & 0.812 ($\pm$0.0001) & \text{-} & 196 \\[.5ex]
\cline{2-7}
& DAE (50\% input dropout)  & 0.461 ($\pm$0.0019) & \textless 0.0001 & 0.807 ($\pm$0.0015) & 0.0103 & 400\\[.5ex]
\cline{2-7}
& DpAE (50\% hidden dropout) & 0.491 ($\pm$0.0013) & \textless 0.0001 & 0.815 ($\pm$0.0038) & \textless 0.0001 & 400\\[.5ex]
\cline{2-7}
&  AAE & \textit{-}  & \textit{-} & 0.791 ($\pm$0.041) & \textless 0.0001 & 1000
\\\cline{1-7}
\hline\hline
\end{tabular}}}
\label{table:result4}
\end{table*}

\subsection{Unsupervised Semantic Feature Learning from Text Data}
In this experiment DpAE, NCAE, and $L_1/L_2$-NCSAE are evaluated and compared based on their ability to extract semantic features from text data, and
how they are able to discover the underlined structure in text data. For this purpose, the Reuters-21578 text categorization dataset with $200$ features is utilized to train all the three types of AEs with $20$ hidden nodes. A subset of $500$ examples belonging to categories "grain", "crude", and "money-fx" was extracted from the test set. The experiments were run three times, averaged and recorded. In Fig.~\ref{tsne_reuters_15_compo1}, the 20-dimensional representations of the Reuters data subset using DpAE, NCAE, and $L_1/L_2$-NCSAE are visualized. It can be observed that $L_1/L_2$-NCSAE is able to disentangle the documents into three distinct categories with more linear manifolds than NCAE. In addition, $L_1/L_2$-NCSAE is able to group documents that are closer in the semantic space into the same categories than DpAE that finds it difficult to group the documents into any distinct categories with less overlap.
\subsection{Supervised Learning}
In the last set of experiments, a deep network was constructed using two stacked $L_1/L_2$-NCSAE and a softmax layer for classification to test if the enhanced ability of the network to shatter data into parts and lead to improved classification. Eventually, the entire deep network is fine-tuned to improve the accuracy of the classification. In this set of experiments, the performance of pre-training a deep network with $L_1/L_2$-NCSAE is compared with those pre-trained with recent AE architectures. The MNIST and NORB data sets were utilized, and every run of the experiments is repeated ten times and averaged to combat the effect of random initialization. The classification accuracy of the deep network pre-trained with NNSAE \cite{Lemme2012OnlineLearning}, DpAE \cite{hinton2012improving}, DAE \cite{vincent2008extracting}, AAE \cite{makhzani2015adversarial}, NCAE, and $L_1/L_2$-NCSAE using MNIST and NORB data respectively are detailed in Table~\ref{table:result4}. The network architectures are 784-196-20-10 and 1024-200-20-5 for MNIST and NORB dataset respectively. It is remarked that for training of AAE with two layers of 196 hidden units in the encoder, decoder, discriminator, and other hyperparameters tuned as described in \cite{makhzani2015adversarial}, the accuracy was $83.67$\%. The AAE reported in Table~\ref{table:result4} used encoder, decoder, and discriminator each with two layers of 1000 hidden units and trained for 1000 epochs. The classification accuracy and speed of convergence are the figures of merit used to benchmark $L_1/L_2$-NCSAE with other AEs.\\
\indent
It is observed from the result that $L_1/L_2$-NCSAE-based deep network gives an improved accuracy before fine-tuning compared to methods such as NNSAE, NCAE, DpAE, and NCAE. However, the performance in terms of classification accuracy after fine-tuning is very competitive. In fact, it can be inferred from the p-value of the experiments conducted on MNIST and NORB in Table~\ref{table:result4} that there is no significant difference in the accuracy after fine-tuning between NCAE and $L_1/L_2$-NCSAE even though most of the weights in $L_1/L_2$-NCSAE are nonnegativity constrained. Therefore it is remarked that even though the interpretability of the deep network has been fostered by constraining most of the weights to be nonnegative and sparse, nothing significant has been lost in terms of accuracy. In addition, network trained with $L_1/L_2$-NCSAE was also observed to converge faster than its counterparts. On the other hand, NNSAE also has nonnegative weights but with deterioration in accuracy, which is more conspicuous especially before the fine-tuning stage. The improved accuracy before fine-tuning in $L_1/L_2$-NCSAE based network can be traced to its ability to decompose data more into distinguishable parts. Although the performance of $L_1/L_2$-NCSAE after fine-tuning is similar to those of DAE and NCAE but better than NNSAE, DpAE, and AAE, $L_1/L_2$-NCSAE constrains most of the weights to be nonnegative and sparse to foster transparency than for other AEs. However, DpAE and NCAE performed slightly more accurate than $L_1/L_2$-NCSAE on NORB after network fine-tuning. \\
\indent
In light of constructing an interpretable deep network, an $L_1/L_2$-NCSAE pre-trained deep network with $10$ hidden neurons in the first AE layer, $5$ hidden neurons in the second AE, and 10 output neurons (one for each category) in the softmax layer was constructed. It was trained on Reuters data, and compared with that pre-trained using DpAE. The interpretation of the encoding layer of the first AE is provided by listing words associated with $10$ strongest weights, and the interpretation of the encoding layer of the second AE is portrayed as images characterized by both the magnitude and sign of the weights. Compared to the AE with weights of both signs shown in Fig.~\ref{reuters}a, Fig.~\ref{reuters}b allows for much better insight into the categorization of the topics. \\
\indent
Topic \emph{earn} in the output weight matrix resonates with the 5th hidden neuron most, lesser with the 3rd, and somewhat with the 4th. This resonance can happen only when the 5th hidden neuron reacts to input by words of columns 1 and 4, and in addition, to a lesser degree, when the 3rd hidden neuron reacts to input by words of the 3rd column of words. So, in tandem, the dominant columns 1, 4 and then also 3 are sets of words that trigger the category \emph{earn}. \\
\indent
Analysis of the term words for the topic \emph{acq} leads to a similar conclusion. This topic also resonates with the two dominant hidden neurons 5 and 3 and somewhat also with neuron 2. These neurons 5 and 3 are driven again by the columns of words 1,4, and 3. The difference between the categories is now that to a lesser degree, the category \emph{acq} is influenced by the 6th column of words. An interesting point is in contribution of the 3rd column of words. The column connects only to the 4th hidden neuron but weights from this neuron in the output layer are smaller and hence less significant than for any other of the five neurons (or rows) of the output weight matrix. Hence this column is of least relevance in the topical categorization.
\subsection{Experiment Running Times}
The training time for networks with and without the nonnegativity constraints was compared. The constrained network converges faster and requires lesser number of training epochs. In addition, the unconstrained network requires more time per epoch than the constrained one. The running time experiments were performed using full MNIST benchmark dataset on Intel(r) Core(TM) i7-6700 CPU @ 3.40Ghz and a 64GB of RAM running a 64-bit Windows 10 Enterprise edition. The software implementation has been with MATLAB 2015b with batch Gradient Descent method, and LBFGS in minFunc (\cite{Byrd1995}) is used to minimize the objective function. The usage times for constrained and unconstrained networks were also compared. We consider the usage time in milliseconds (ms) as the time elapsed in ms a fully trained deep network requires to classify all the test samples. The unconstrained network took 48 ms per epoch in the training phase while the constrained counterpart took 46 ms. Also, the unconstrained network required 59.9 ms usage time, whereas the network with nonnegative weights took 55 ms. From the above observations, it is remarked that the nonnegativity constraint simplifies the resulting network.

\section{Conclusion}					
This paper addresses the concept and properties of special regularization of DL AE that takes advantage of non-negative encodings and at the same time of special regularization. It has been shown that by using both $L_1$ and $L_2$ to penalize the negative weights, most of them are forced to be nonnegative and sparse, and hence the network interpretability is enhanced. In fact, it is also observed that most of the weights in the Softmax layer become nonnegative and sparse. In sum, it has been observed that encouraging nonnegativity in NCAE-based deep architecture forces the layers to learn part-based representation of their input and leads to a comparable classification accuracy before fine-tuning the entire deep network and not-so-significant accuracy deterioration after fine-tuning. It has also been shown on select examples that concurrent $L_1$ and $L_2$ regularization improve the network interpretability. The performance of the proposed method was compared in terms of sparsity, reconstruction error, and classification accuracy with the conventional SAE and NCAE, and we utilized MNIST handwritten digits, Reuters documents, and the NORB dataset to illustrate the proposed concepts.

\bibliographystyle{IEEEtran}
\bibliography{autoencoder}
\vspace{-15mm}
\begin{IEEEbiography}[{\includegraphics[width=1in,height=1.25in,clip,keepaspectratio]{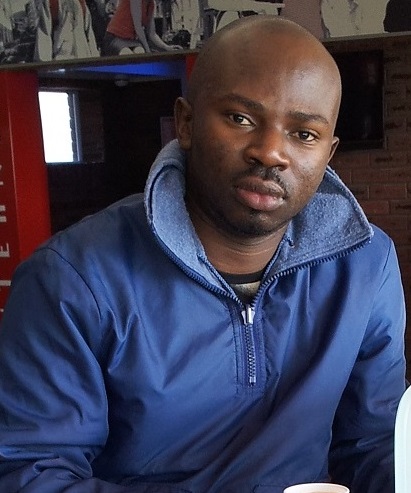}}]
{Babajide Ayinde} (S'09) received the M.Sc. degree in Engineering Systems and Control from the King Fahd University of Petroleum and Minerals, Dhahran, Saudi Arabia. He is currently a Ph.D. student at the University of Louisville, Kentucky, USA and a recipient of University of Louisville fellowship. His current research interests include unsupervised feature learning and deep learning techniques and applications.
\end{IEEEbiography}
\begin{IEEEbiography}[{\includegraphics[width=1in,height=1.25in,clip,keepaspectratio]{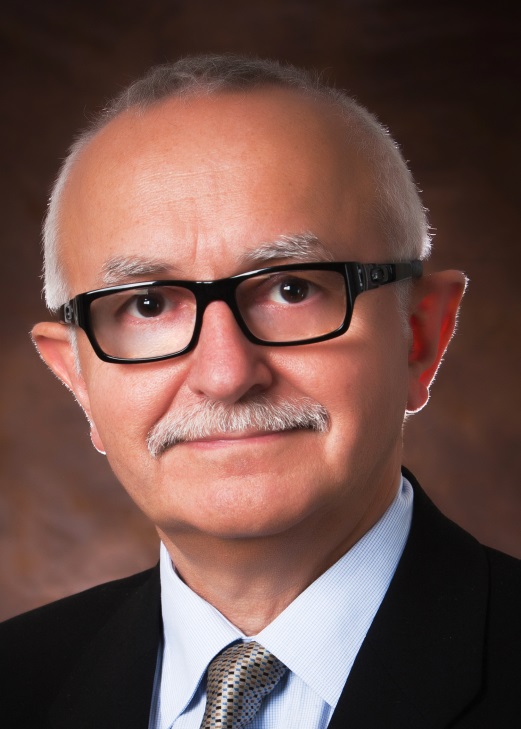}}]
{Jacek M. Zurada}
(M'82-SM'83-F'96-LF'14) Ph.D., has received his degrees from Gdansk Institute of Technology, Poland. He now serves as a Professor of Electrical and Computer Engineering at the University of Louisville, KY. He authored or co-authored several books and over 380 papers in computational intelligence, neural networks, machine learning, logic rule extraction, and bioinformatics, and delivered over 100 presentations throughout the world.

In 2014 he served as IEEE V-President, Technical Activities (TAB Chair). He also chaired the IEEE TAB Periodicals Committee, and TAB Periodicals Review and Advisory Committee and was the Editor-in-Chief of the IEEE Transactions on Neural Networks (1997-03), Associate Editor of the IEEE Transactions on Circuits and Systems, Neural Networks and of The Proceedings of the IEEE. In 2004-05, he was the President of the IEEE Computational Intelligence Society.

Dr. Zurada is an Associate Editor of Neurocomputing, and of several other journals. He has been awarded numerous distinctions, including the 2013 Joe Desch Innovation Award, 2015 Distinguished Service Award, and five honorary professorships.  He has been a Board Member of IEEE, IEEE CIS and IJCNN.
\end{IEEEbiography}
\end{document}